\documentclass{article}

\usepackage[hyphens]{url}
\usepackage{scalerel}
\usepackage{natbib}  % DO NOT CHANGE THIS AND DO NOT ADD ANY OPTIONS TO IT
\usepackage{bibentry}
\setcitestyle{numbers}
\usepackage{hyperref}       % hyperlinks
\usepackage{url}          % simple URL typesetting
\setcitestyle{square}
    \usepackage[final]{neurips_2021}

\usepackage[utf8]{inputenc} % allow utf-8 input
\usepackage[T1]{fontenc}    % use 8-bit T1 fonts
\usepackage{hyperref}       % hyperlinks
\usepackage{url}            % simple URL typesetting
\usepackage{booktabs}       % professional-quality tables
\usepackage{amsfonts}       % blackboard math symbols
\usepackage{nicefrac}       % compact symbols for 1/2, etc.
\usepackage{microtype}      % microtypography
\usepackage{xcolor}         % colors
\usepackage{algorithm}
\usepackage{algorithmic}

\usepackage[english]{babel}
\usepackage{amsthm}
\usepackage{tikz}
\theoremstyle{remark}

\title{Sign-to-Speech Model for Sign Language Understanding: A Case Study of Nigerian Sign Language}

\author{%
  Steven Kolawole \\
  Federal University of Agriculture, Abeokuta\\
  ML Collective\\
  \texttt{kolawolesteven99@gmail.com} \\
   \And
   Opeyemi Osakuade \\
    Data Science Nigeria \\
   \texttt{osakuade@datasciencenigeria.ai} \\
   \AND
   Nayan Saxena\\
  University of Toronto \\
  ML Collective\\
  \texttt{nayan.saxena@mail.utoronto.ca}
   \And
   Babatunde Kazeem Olorisade \\
   Cardiff Metropolitan University \\
   \texttt{kolorisade@cardiffmet.ac.uk} \\
}

\begin{document}
\theoremstyle{definition}
\newtheorem{definition}{Definition}
\maketitle

\begin{abstract}
Through this paper we seek to reduce the communication barrier between the hearing-impaired community and the larger society who are usually not familiar with sign language in the sub-Saharan region of Africa with the largest occurrences of hearing disability cases, while using Nigeria as a case study. The dataset is a pioneer dataset for the Nigerian Sign Language and was created in collaboration with relevant stakeholders. We pre-processed the data in readiness for two different object detection models and a classification model and employed diverse evaluation metrics to gauge model performance on sign-language to text conversion tasks. Finally, we convert the predicted sign texts to speech and deploy the best performing model in a lightweight application that works in real-time and achieves impressive results converting sign words/phrases to text and subsequently, into speech. 
\end{abstract}

\section{Introduction}
Communication has always been the mainstay of functional human interaction in any society. The hearing-impaired community uses sign language for effective interpersonal communication. Whilst the use of sign language works well as a means of communication between the hearing-impaired community, it is not the same while attempting to communicate with people outside of their community.
In 2012, WHO estimated that over 5.3\% of the world’s population (about a 430 million) have hearing disabilities and it is most prevalent in sub-Saharan Africa \citep{world}. Despite this prevalence, facilities to support their communication with the larger society are still lacking, most especially in developing countries. Children with hearing loss and deafness often do not receive formal schooling; adults with hearing loss suffer a higher unemployment rate;  a higher percentage of those employed are in the lower grades of employment compared with the general workforce. There are also other impacts including social isolation, loneliness, and stigmatization.
Although there has been an evolution of computer vision with artificial intelligence in creating innovations to solve hearing disabilities problems, we have very few of these solutions targeted towards developing countries \cite{inclusive}. This is mostly due to two factors: (i) The sign language data in the region is low resourced  (ii) There are increasing complexities and advanced tools required to deploy these solutions in real-life environments.

% \section{Literature review}
\paragraph{Related Work} Dataset is one of the common needs that could blend the recognition,translation,and generation technologies for sign language since modern, data-driven machine learning techniques work best in data-rich scenarios \cite{bragg2019sign}. There is an abundance of American Sign Language datasets available publicly as a high-resource sign language, with the most recent one being a large-scale Word-Level American Sign Language dataset \cite{li2020word}. Sign languages just like spoken languages vary considerably from one location to another. INCLUDE \cite{sridhar2020include}, an Indian Sign Language dataset contains 0.27 million frames across 4,287 videos over 263-word signs from 15 different word categories, but  when it comes to the region with the highest number of hearing-impaired population, i.e. sub-Saharan Africa, there is a dearth of sign language datasets. Aside from an instance of a slightly-complex South African Sign Language dataset which uses kinetic gloves \cite{mcinnes2014south} and a very small corpus of Ghanaian Sign Language which serves as a proof of concept \cite{odartey2019ghanaian}, no further work has been done in creating standard datasets for sub-Saharan African sign languages. 

In this research project, we facilitate the creation of low-resource sign language datasets for countries where hearing impairments are most prevalent using the Nigeria Sign Language as a case study. A good dataset should sufficiently represent a challenging problem to make it both useful and to ensure its longevity -- to the authors knowledge this dataset is the first of its kind for the low-resource sign languages in sub-Saharan Africa. The dataset images were annotated for object detection using LabelImg \cite{lbl}, in both YOLOv and PASCAL VOC formats. Furthermore, two object detection models and a classification model were trained and compared across multiple evaluation metrics. 
The results of this work clearly demonstrate that if provided the availability of otherwise low-resource data, the communication barrier between the hearing impaired community and the larger society can be bridged using the sign-to-speech machine learning techniques that can further be deployed to work in real-time.

\section{ Nigerian Sign Language Dataset}
The dataset comprises of around 5000 images with 137 sign words, including the 27 alphabet letters. It should be noted that sign words and phrases are not just dependent on the signs being performed, but on the facial expression, the body posture, the relative position of the signs to the body, and motion. In fact, some signs are disambiguated based only on motion making it challenging to create a dataset that is non-complex and easy to reuse. Hence, for signs that are heavily dependent on motion, we tried to capture the point in the motion where that moment is peculiar to the specific sign only. Supplementary materials, including code and model configs, are made available on Github {{\href{https://github.com/SteveKola/Sign-to-Speech-for-Sign-Language-Understanding}{\color{blue}here}}}.

\begin{figure*}[htbp]
\centering
\includegraphics[width=0.6 \textwidth]{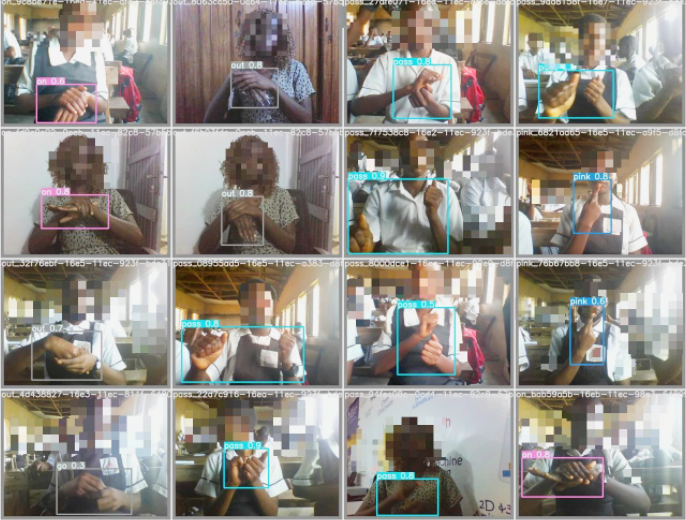}
\caption{{A mosaic of a fraction of the dataset in diverse backgrounds and lighting conditions.}}
\end{figure*}

\subsection{Dataset Creation}
The dataset focuses on the Nigerian Sign Language and the initial data was created by a TV sign language broadcaster from the Ogun State Broadcasting Corporation. Further works to expand the dataset, with the aim of creating a wider dispersion in the dataset’s distribution, were carried out in conjunction with teachers and students totalling 20 individuals from two special education schools in Abeokuta and Lagos-- both cities in Nigeria. Figure 1 reflects the diverse backgrounds and lighting conditions in which the images were captured.

The Nigerian Sign Language, like most of the other sign languages in the world, is quite influenced by the American Sign language, with a chunk of the language adopted from the British Sign Language and a smattering of it being vernacular signs. 
As shown in Figure 1, an initial 8000 static images of 137 signs words/phrases, including the 27 alphabet letters, were created with 20+ individuals in diverse environments, under different environments, to account for the different environments in which the work might be applied.

\subsection{Preprocessing \& Annotation}

Data cleaning was performed to weed out excessively blurry images and images where the signs were not totally contained in the image frames. This reduced our images from 8000 to 5000 images. All images were resized into 640 x 640, therefore having a shape of (640, 640, 3). Images of the same class names were stored in the same image self-named folders formatted as “\texttt{classname\_id.jpg}”.

A core part of computer vision tasks is to label the data, making it essential for the dataset  to be annotated before it can be used with object detection models. This was done using LabelImg$^2$, a graphical image labeling tool, which was used to draw bounding boxes around the signs performed in each image. These rectangular bounding boxes  define the location of the target object (sign in this case) with each bounding box consisting of the $x$ and $y$ coordinates (xmin-top left, ymin-top left, xmax-bottom right, ymax-bottom right) \cite{everingham2010pascal}.
The annotations were created in both PASCAL VOC format and the YOLO format. PASCAL VOC stores the annotations as \texttt{.xml} files while the YOLO labeling format stores  \texttt{.txt} files.

\section{Modelling Experiments and Evaluation}

We designed three models using two object detection architectures, YOLO and Single-Shot-Detector, and a fine-tuned pre-trained model for classification and compared results across the three models.

\paragraph{Evaluation Metrics} The Precision is calculated as the ratio between the number of positive samples correctly classified to the total number of samples classified as positive (either correctly or incorrectly) while the recall is calculated as the ratio between the number of positive samples correctly classified as positive to the total number of positive samples. The recall measures the model's ability to detect positive samples and the precision measures the model's accuracy in classifying a sample as positive. A precision-recall (PR) curve shows the trade-off between the precision and recall values for different thresholds and the mAP is a way to summarize the PR-curve into a single value representing the average of all precisions.  Training an object detection model usually requires two inputs which are the image and the ground-truth bounding boxes for the object in each image. When the model predicts the bounding box, it is expected that the predicted box will not match exactly the ground-truth box. IOU (Intersection over Union) is calculated by dividing the area of intersection between the two boxes by the area of their union which means that higher IOU scores generally translate into better predictions. Specifically, in this paper, we measure where there is a 50\% overlap (@0.5) and where there is a 95\% overlap (@0.95).

\subsection{Object Detection Using YOLO}

YOLO’s unified architecture is an extremly fast architecture that reframes object detection as a single regression problem, straight from image pixels to bounding box coordinates and class probabilities. Using this architecture, you only look once (YOLO) at an image to predict what objects are present and where they are \cite{redmon2016you}.
We made use of YOLOv5m implementation with the annotations in YOLO format and trained across 150 epochs. Data augmentation techniques including scaling, left-right flipping, HSV (Hue, Saturation, and Value) manipulation among others, were performed on the data.
After the training process, we evaluated the performance of the model using several metrics including Precision, Recall, and mAP (mean Average Precision) when IOU are at 0.5 (50\%) and 0.95 (95\%). Figure 2 shows the graphs of the metrics curves as training progresses.
After evaluation, the YOLO model had a validation precision score of 0.8057, recall score of 0.95, as well as mAP scores of 0.95 and 0.64 for @0.5IOU and @0.95IOU respectively. This result confirms the effectiveness of our approach in predicting signs performed in diverse environments correctly.

\begin{figure*}[ht]
\centering
\includegraphics[width=0.8 \textwidth]{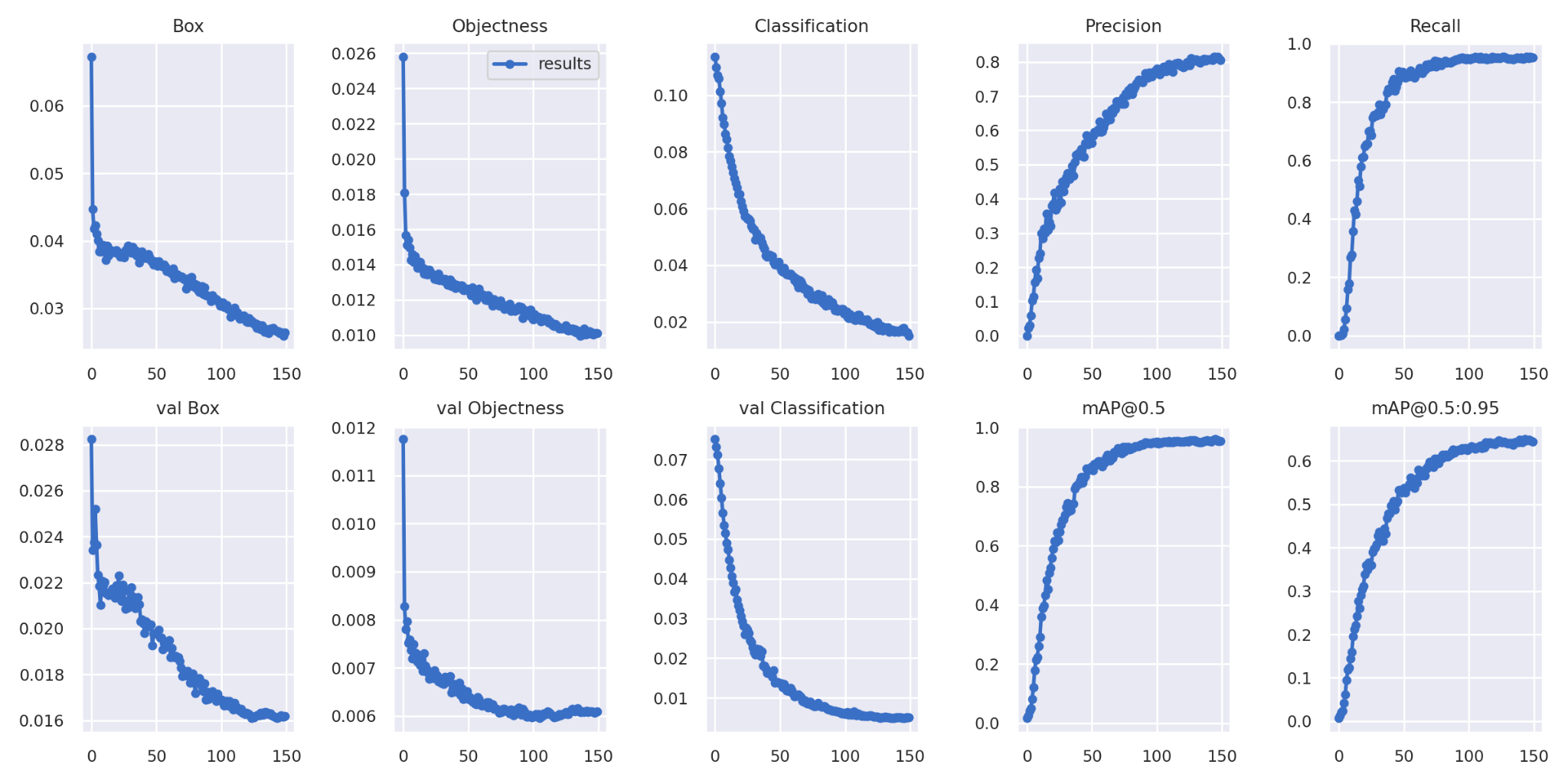}
\caption{{Graph of Precision, Recall, and mAP as YOLOv5 training progresses.}}
\end{figure*}

\subsection{Object Detection using Single-Shot Detector}
The Single Shot Detector (SSD) architecture uses a single deep neural network in predicting category scores and box offsets for a fixed set and  achieves high detection accuracy by producing predictions of different scales from feature maps of different scales and explicitly separate predictions by aspect ratio \cite{liu2016ssd}.
We used TensorFlow Object Detection API to access and finetune the SSD ResNet50 V1 FPN model, which was pre-trained on the COCO 2017 dataset \cite{lin2014microsoft}, and we fed the model our dataset which was converted into TensorFlow Record format. We made use of only horizontal flip and image cropping for Data augmentation, and we train across 40,000 train steps. Then we evaluate the model on the test set using Precision, Recall, and mAP metrics. After evaluation, the SSD model achieved a Precision score of 0.6414, a Recall score of 0.7075, mAP scores of 0.9535 and 0.6412 for 50\% and 95\% overlap respectively. This result is not as good as the YOLO model’s result. Further investigations revealed that the SSD model is a little bit “fond” of imagining signs from random shapes. We deduced that the SSD model might require a much larger dataset to produce results that are on par with the YOLO model’s result.

\subsection{Classification using Transfer Learning}

\begin{figure*}[htbp]
\centering
\includegraphics[width= 0.35\textwidth]{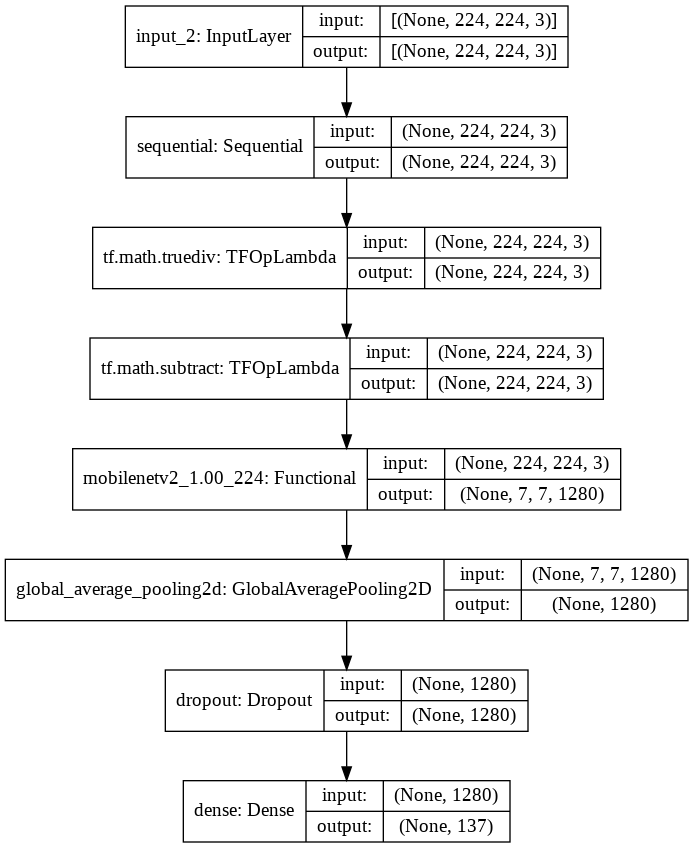}
\caption{{Model architecture using MobileNet V2 as the pre-trained model.}}
\end{figure*}

We decided to model the problem as a classification problem and train on the dataset while discarding the annotations. The images were resized to 224 by 224 pixels and applied to a pre-trained image classification  model known as MobileNet V2 \cite{sandler2018mobilenetv2} which was then customized in 2 ways.

\paragraph{Feature Extraction on a Pretrained Model}
We used the representation learned by MobileNetV2 when pre-trained on the ImageNet dataset \cite{sandler2018mobilenetv2} to extract features from our data. We built a small model atop the pre-trained model without training any layers from the pre-trained model. Upon evaluation after 60 epochs, our model achieved a test accuracy score of 0.4397, precision score of 0.8640, and 0.1401 recall score.

\paragraph{Finetuning on the Pretrained Model}
The poor performance indicated that the model required fine-tuning. MobileNetV2 has 154 layers, hence we left only the first 100 layers frozen and train on both the newly-added classifier layers and the last layers of the pre-trained model. This helped fine-tune the higher-order representations in the base model thereby making it more relevant for our specific task.  We evaluated the model after training for an additional 140 epochs and the results are: 0.9115 accuracy score, 0.9355 precision score, and 0.9063 recall score.

\begin{figure*}[ht]
\centering
\includegraphics[width= 0.7\textwidth]{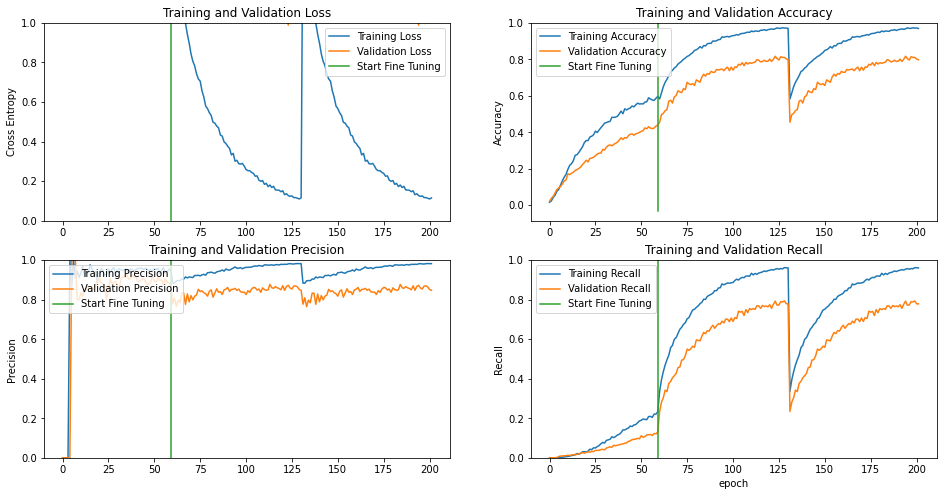}
\caption{Graphs comparing the feature extraction performance with the fine-tuned performance.}
\end{figure*}

By finetuning the pretrained model, we reduced the loss sporadically and both the accuracy score and precision score skyrocketed as seen in Figure 4 and Table 1.

\begin{table}[htbp]
\label{tbl}
  \centering
  \resizebox{0.4\columnwidth}{!}
  {
  \begin{tabular}{|c|c|c|c|}
    \hline
\textsc{Metrics} &  \textsc{YOLO} & \textsc{SSD} & \textsc{Classification} \\
\hline
  Recall & 0.9512 & 0.7075 & 0.9355  \\
  \hline
  Precision & 0.806 & 0.6414 & 0.9063  \\
    \hline
  mAP:@0.5 & 0.9533 & 0.9535& N/A \\
   \hline
  mAP:@0.95 &0.6439&0.6412&N/A\\
    \hline
    \end{tabular}}
    \medskip

     \caption{Comparison of Precision, Recall, and mAP across different models.}
\end{table}
% \section{Real-time Development}

\section{Conclusion} In this paper, we created a dataset for a low-resource sign language and we experimented with different models and deployed the best model for real-time sign-to-speech synthesis. In pursuance of the objective of this paper which is to bridge the communication barrier between the hearing impaired community and the larger society, we enabled text-to-speech synthesis and deployed the YOLO model for real-time usage in production. This is being done by converting the sign texts (or labels) that are being returned by the model into an equivalent voice of words using Pyttsx3, a text-to-speech conversion library in Python, and deploying the model on DeepStack Server for real-time usage. The authors hope that this work sufficiently demonstrates how much the alienation of the hearing impaired community by the larger society in developing countries can be mitigated using resource-efficient machine learning techniques and tools.

% \pagebreak
% \newpage
\section*{Acknowledgements} Many thanks to Amanda Bibire of Ogun State Broadcasting Corporation, for volunteering to create the first batch of the dataset. We appreciate the teachers and students of the Special Education School, Saint Peter's College, Abeokuta, for graciously dedicating 4 class sessions to create the later and larger batch of the dataset. Finally, we thank the ML Collective community for the generous computational support, as well as helpful discussions, ideas, and feedback on experiments.

\bibliography{manuscript}

% \section*{Appendix}

% \subsection*{A. Hyperparameters}

% \subsection*{B. }

\end{document}